\documentclass{article}

\usepackage{microtype}
\usepackage{graphicx}
\usepackage{subcaption}
\usepackage{booktabs} %

\usepackage{hyperref}

\usepackage[preprint]{icml2026}

\usepackage{amsmath}
\usepackage{amssymb}
\usepackage{mathtools}
\usepackage{amsthm}

\usepackage{amsmath,amsfonts,bm}

\def\1{\bm{1}}

\DeclareMathAlphabet{\mathsfit}{\encodingdefault}{\sfdefault}{m}{sl}
\SetMathAlphabet{\mathsfit}{bold}{\encodingdefault}{\sfdefault}{bx}{n}

\DeclareMathOperator*{\argmax}{arg\,max}

\usepackage{times}
\usepackage{latexsym}
\usepackage{balance}

\usepackage[T1]{fontenc}

\usepackage[utf8]{inputenc}

\usepackage{inconsolata}

\usepackage{url}
\usepackage{bm}
\usepackage{xspace}
\definecolor{darkblue}{rgb}{0, 0, 0.5}
\hypersetup{colorlinks=true, citecolor=darkblue, linkcolor=darkblue, urlcolor=darkblue}
\usepackage{adjustbox}
\usepackage{multirow}
\usepackage{algorithm}
\usepackage{algorithmic}

\usepackage{amsfonts}
\usepackage{multicol}

\usepackage{xcolor}
\usepackage{colortbl}

\usepackage{array}

\usepackage{here}

\usepackage{verbatim}
\usepackage{tcolorbox}
\usepackage{listings}

\tcbuselibrary{breakable,skins}
\lstdefinestyle{prompttemplate}{
  basicstyle=\ttfamily\footnotesize,
  columns=fullflexible,
  breaklines=true,
  breakatwhitespace=false,
  keepspaces=true,
  showstringspaces=false,
  frame=none,
  aboveskip=0pt,
  belowskip=0pt,
}
\newtcolorbox{promptbox}[1]{
  enhanced,
  breakable,
  colback=gray!3,
  colframe=gray!45,
  boxrule=0.4pt,
  arc=1.2mm,
  left=2mm,
  right=2mm,
  top=1mm,
  bottom=1mm,
  colbacktitle=gray!12,
  coltitle=black,
  fonttitle=\bfseries,
  title={#1},
}

\usepackage{bbm} %

\newcommand{\ind}{\mathbbm{1}} %
\newcommand{\Ind}[1]{\ind\!\left[#1\right]} %

\usepackage[capitalize,noabbrev]{cleveref}

\theoremstyle{plain}

\theoremstyle{definition}

\theoremstyle{remark}

\usepackage[textsize=tiny]{todonotes}

\icmltitlerunning{Where-to-Unmask: Ground-Truth-Guided Unmasking Order Learning for Masked Diffusion Language Models}

\begin{document}

\twocolumn[
  \icmltitle{Where-to-Unmask: Ground-Truth-Guided \\ Unmasking Order Learning for Masked Diffusion Language Models}

  \icmlsetsymbol{equal}{*}

  \begin{icmlauthorlist}
    \icmlauthor{Hikaru Asano}{utokyo,riken}
    \icmlauthor{Tadashi Kozuno}{osx}
    \icmlauthor{Kuniaki Saito}{osx}
    \icmlauthor{Yukino Baba}{utokyo}
  \end{icmlauthorlist}

  \icmlaffiliation{utokyo}{The University of Tokyo, Tokyo, Japan}
  \icmlaffiliation{riken}{RIKEN AIP, Tokyo, Japan}
  \icmlaffiliation{osx}{OMRON SINIC X, Tokyo, Japan}

  \icmlcorrespondingauthor{Yukino Baba}{yukino-baba@g.ecc.u-tokyo.ac.jp}

  \icmlkeywords{Machine Learning, ICML}

  \vskip 0.3in
]

\printAffiliationsAndNotice{}  %

\begin{abstract}
  Masked Diffusion Language Models (MDLMs) generate text by iteratively filling masked tokens, requiring two coupled decisions at each step: which positions to unmask (where-to-unmask) and which tokens to place (what-to-unmask).
  While standard MDLM training directly optimizes token prediction (what-to-unmask), inference-time unmasking orders (where-to-unmask) are typically determined by heuristic confidence measures or trained through reinforcement learning with costly on-policy rollouts.
  To address this, we introduce Gt-Margin, a position-wise score derived from ground-truth tokens, defined as the probability margin between the correct token and its strongest alternative.
  Gt-Margin yields an oracle unmasking order that prioritizes easier positions first under each partially masked state.
  We demonstrate that leveraging this oracle unmasking order significantly enhances final generation quality, particularly on logical reasoning benchmarks.
  Building on this insight, we train a supervised unmasking planner via learning-to-rank to imitate the oracle ordering from masked contexts.
  The resulting planner integrates into standard MDLM sampling to select where-to-unmask, improving reasoning accuracy without modifying the token prediction model.
\end{abstract}

\begin{figure}[t]
    \centering
    \includegraphics[width=\linewidth]{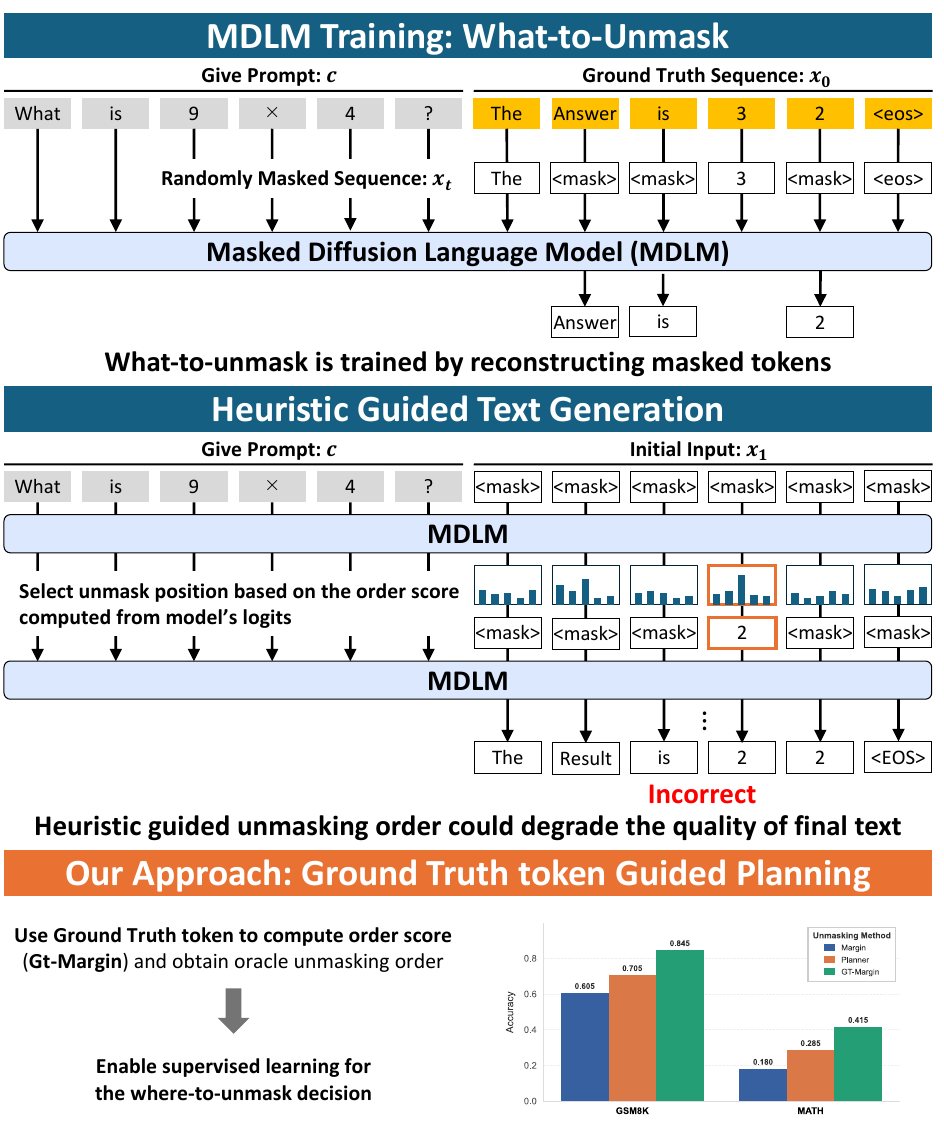}
    \caption{MDLMs learn \emph{what-to-unmask} but leave \emph{where-to-unmask} implicit, and heuristic scores (e.g., Margin) can yield incorrect outputs. We define Gt-Margin using ground-truth tokens and train a planner to imitate the oracle ordering, improving reasoning accuracy without modifying the token model.}
    \label{fig:teaser}
\end{figure}

\section{Introduction}

Masked Diffusion Language Models (MDLMs)~\citep{Li2025-rv, Nie2025-ne,Ye2025-pr,Bie2025-ae} have recently emerged as a powerful alternative to autoregressive (AR) models by formulating text generation as a diffusion-based denoising process~\citep{Austin2021-pj,Campbell2022-xd,Lou2024-op,Shi2024-mq, Sahoo2025-tm}.
Instead of being constrained to left-to-right generation, MDLMs iteratively reconstruct tokens masked by a special token (\texttt{<mask>}) in arbitrary order~\citep{kim2025train}.
This flexibility naturally benefits non-monotonic tasks such as infilling and editing~\citep{Ghazvininejad2019-ec,Stern2019-xw,Lee2025-qu,Zhang2025-cx}.
Recent studies further demonstrate that MDLMs achieve superior performance on logical reasoning tasks, including programming and mathematics~\citep{Ye2024-md,Gong2025-ta,Svete2025-fl}, and can learn more effectively even from limited data~\citep{Ni2025-lf}.

Text generation with MDLMs involves two decisions at each step: determining \textbf{which position to unmask next} (\textit{where-to-unmask}) and identifying \textbf{which token to predict at that position} (\textit{what-to-unmask}).
While \textit{what-to-unmask} determines which token to place at a chosen position by sampling from the vocabulary distribution,
\textit{where-to-unmask} determines which position gets unmasked next, thereby controlling what information is revealed early and what context becomes available to condition all subsequent predictions~\citep{Lee2025-up,Peng2025-oi}.
Empirical evidence consistently highlights the crucial role of the unmasking order in determining the quality of the final generated sequence~\citep{Ben-Hamu2025-fi,Hong2025-np,kim2025train}.

Despite its importance, standard MDLM training optimizes a supervised likelihood objective that reconstructs randomly masked tokens, explicitly addressing \textit{what-to-unmask} but leaving \textit{where-to-unmask} largely implicit~\citep{Shi2024-mq,Lou2024-op,Sahoo2025-tm}.
Consequently, inference-time unmasking orders rely on heuristics driven by model uncertainty~\citep{Chang2022-xf,Koh2025-qy,kim2025train}.
Recent RL methods introduce token-level rewards~\citep{Gong2025-ta,Ou2025-oh,Wang2025-ql,Zhao2025-in} or explicit planners~\citep{Hong2025-np} to optimize where-to-unmask, but require costly on-policy rollouts.
Thus, an efficient \textbf{supervised framework that directly leverages ground-truth sequences for training the \textit{where-to-unmask} decision remains an open challenge}.

In this work, we address this gap by proposing a straightforward question: \textbf{if we had access to the ground-truth completion, what unmasking order would be optimal?}
We propose \texttt{Gt-Margin}, a position-wise metric that quantifies how decisively the model prefers the ground-truth token at each masked position.
Concretely, at a generation state, Gt-Margin at position $i$ is defined as the gap between the model probability assigned to the ground-truth token and the highest probability among all other tokens.
This metric induces an \emph{oracle} unmasking order by ranking positions from easiest (large margin) to hardest (small margin), explicitly highlighting which positions become clearer and easier to resolve as more context becomes available.

Empirically, we show that ground-truth-derived supervision \emph{solely for the unmasking order} significantly improves performance on logical reasoning tasks without increasing inference cost or modifying the token prediction model.

Building on this oracle, we train a supervised \emph{unmasking planner} to predict the oracle-induced ranking from partially masked sequences.
We formulate this as a learning-to-rank problem~\citep{Liu2009-mf,Swezey2021-fg}, where the planner scores masked positions to match the Gt-Margin-derived oracle ordering.
This planner integrates seamlessly into standard MDLM sampling to determine \emph{where-to-unmask} during inference, without modifying the token prediction model.
Experimental results confirm that our learned planner generalizes well and outperforms heuristic baselines.

Our contributions are:
(i) We define \textbf{Gt-Margin} as an oracle measure that induces an easy-to-hard unmasking ordering.
(ii) We analyze the \textbf{effect of oracle ordering}, revealing when early steps matter most.
(iii) We train a \textbf{deployable unmasking planner} using oracle supervision, enabling improved sampling \textbf{without modifying the base MDLM}.

\section{Preliminaries}
\label{sec:prelim}

We consider conditional generation of a length-$L$ sequence $\mathbf{x}=(x_1,\ldots,x_L)\in\mathcal{V}^L$ over a vocabulary $\mathcal{V}$, optionally conditioned on a prompt $\mathbf{c}$. The prompt is always observed and never masked. Let $m=\texttt{<mask>}$ denote the absorbing mask token, and let $\tilde{\mathcal{V}}=\mathcal{V}\cup\{m\}$ be the extended vocabulary.
At time $t\in[0,1]$, the generation state is a partially observed sequence $\mathbf{x}_t\in\tilde{\mathcal{V}}^L$, where $\mathbf{x}_{t}^{\,i}=m$ indicates that the token at position $i$ is masked.
Throughout, superscripts index sequence positions (e.g., $i$ in $\mathbf{x}_t^{\,i}$), while subscripts index diffusion time (e.g., $t$ in $\mathbf{x}_t^{\,i}$).
Let $\delta_a$ denote the point mass at $a$, i.e., $\delta_a(z)=\Ind{z=a}$.
Although we use continuous time $t\in[0,1]$ in the exposition, we often discretize it as $t_k=k/T$ for $k=0,\ldots,T$ and write $\mathbf{x}_k:=\mathbf{x}_{t_k}$.

\subsection{Masked Diffusion for Language}

\paragraph{Forward process (masking).}
MDLMs define a forward process that incrementally masks a clean sequence $\mathbf{x}_0 = (x_0^1,\ldots,x_0^L)$.
We use a masking schedule $\alpha_t\in[0,1]$, the probability that a token is \emph{unmasked} at time $t$, with $\alpha_0=1$ and $\alpha_1=0$.
Under position-wise independent masking, the marginal corruption distribution is
\begin{equation}
q(\mathbf{x}_t\,|\,\mathbf{x}_0)
=\prod_{i=1}^{L}\Big(\alpha_t\,\delta_{x_{0}^{i}}(\mathbf{x}_{t}^{i})+(1-\alpha_t)\,\delta_{m}(\mathbf{x}_t^{\,i})\Big).
\end{equation}

Equivalently, on the discrete grid one may parameterize the schedule via step noises $\{\beta_k\}_{k=1}^T$ where each unmasked token is replaced by $m$ with probability $\beta_k$ at step $k$. The corresponding unmasked probability is
\begin{equation}
\alpha_k := \prod_{j=1}^{k}(1-\beta_j),
\end{equation}
and $m$ is \emph{absorbing}, once masked, the token remains masked in the forward process.

\paragraph{Reverse process (generation).}
The reverse (generation) process starts from the fully masked state $\mathbf{x}_T=(m,\ldots,m)$ and iteratively unmasks tokens to recover $\mathbf{x}_0$, proceeding in reverse time $t:1\to0$ using a prediction model $\mu_\theta(\mathbf{x}_t,\mathbf{c})$, typically parameterized as a neural network.

For a reverse-time update from $t$ to $s$ with $0\le s<t\le 1$, let
$M_t := \{\, i \in [L] \mid \mathbf{x}_t^{\,i}=m \,\}$ denote the set of masked positions at time $t$.
Each currently masked position is revealed with probability $\pi_{t\to s} := \frac{\alpha_s-\alpha_t}{1-\alpha_t}$.

Given this unmasking amount, we next decide which positions to unmask (where-to-unmask).
Theoretically, this is done by drawing $u_i \sim \mathrm{Bernoulli}(\pi_{t\to s})$ independently for each $i\in M_t$ and setting $U_{t\to s}:=\{\, i\in M_t \mid u_i=1 \,\}$ as the set of positions to unmask.
In practice, however, $U_{t\to s}$ is often selected by a heuristic metric (Sec.~\ref{sec:where_to_unmask}).
For each $i\in U_{t\to s}$, we sample $x_s^{\,i}\sim \mu_{\theta}^i(\cdot\mid \mathbf{x}_t,\mathbf{c})$, while all other positions remain unchanged. The per-position transition is
\begin{equation}
p_\theta(x_s^{\,i}\,|\,\mathbf{x}_t,\mathbf{c})=
\begin{cases}
\mu_{\theta}^i(x_s^{\,i}\,|\,\mathbf{x}_t,\mathbf{c}) & (\mathbf{x}_t^{\,i}=m,\ i\in U_{t\to s}),\\
\delta_{x_t^{\,i}}(x_s^{\,i}) & (\text{otherwise}). \nonumber
\end{cases}
\end{equation}
Intuitively, visible tokens ($\mathbf{x}_t^{\,i}\neq m$) are deterministically preserved, while masked tokens are either kept masked ($i\notin U_{t\to s}$) or unmasked by sampling ($i\in U_{t\to s}$).

\paragraph{Training objective.}
Standard MDLM training optimizes a supervised likelihood objective that reconstructs randomly masked tokens~\citep{Lou2024-op,Sahoo2025-tm,Shi2024-mq}.
Specifically, we sample $t\sim \mathrm{Unif}(0,1)$ and draw $\mathbf{x}_t\sim q(\cdot\mid \mathbf{x}_0)$. The learning objective minimizes a weighted negative log-likelihood over masked positions:
\begin{equation}
\mathcal{L}(\theta) = - \mathbb{E}_{t,\mathbf{x}_0,\mathbf{x}_t}\left[
\frac{\alpha_t'}{1-\alpha_t}
\sum_{i:\mathbf{x}_t^{\,i}=m} \log \mu_{\theta}^i(x_0^{\,i}\,|\,\mathbf{x}_t,\mathbf{c})
\right], \nonumber
\end{equation}
where $\alpha_t'$ denotes the time derivative of $\alpha_t$. This objective directly optimizes token prediction (what-to-unmask), but leaves the unmasking order (where-to-unmask) implicit.

\subsection{Learning-to-Rank View of Unmasking Order}
\label{subsec:ltr_unmask}

\paragraph{Unmasking order as learning to rank}
At each reverse-time step, the model must decide \emph{where to unmask} among the currently masked positions $M_t := \{\, i \in [L] \mid \mathbf{x}_t^{\,i}=m \,\}$, with $M:=|M_t|$.
We view this decision as a learning-to-rank (LTR) problem: given a query context and a set of items,
an LTR model assigns scores that induce a ranking by priority~\citep{Burges2005-pa,Liu2009-mf,Lee2024-qx}.

Let $Q$ denote the query context space and $Z$ the item feature space.
The query is $q := (\mathbf{c}, \mathbf{x}_t) \in Q$, the prompt and partially masked sequence.
Each item corresponds to a masked position $i\in M_t$ with feature vector $\mathbf{z}_i \in Z$ (e.g., the hidden representation at position $i$).
We assume \emph{ground-truth relevance labels} $\mathbf{y}=\{y_i\}_{i\in M_t}\in\mathbb{R}^{M}$,
where larger $y_i$ indicates higher unmasking priority.
A \emph{planner} learns a scoring function
$f_\phi: Q \times Z^{M} \to \mathbb{R}^{M}$ that outputs scores
$\mathbf{s}=f_\phi(q,\{\mathbf{z}_i\}_{i\in M_t})$.
The planner is trained so that the ranking induced by $\mathbf{s}$ matches the target ordering from $\mathbf{y}$.
Sorting $\mathbf{s}$ in descending order yields a ranking where \emph{rank 1 has highest priority} (unmasked earliest).

\paragraph{Listwise LTR objective for the planner}
Classic LTR objectives are often categorized as pointwise (regressing per-item relevance)~\citep{Cossock2006-dt,Li2007-vq} or pairwise (learning relative orderings)~\citep{Cao2006-gr,Zheng2007-uv}.
Recent work adopts listwise objectives to leverage the ground-truth relevance labels $\mathbf{y}$~\citep{Taylor2008-bh,Xia2008-og,Swezey2021-fg}.

A standard listwise metric is discounted cumulative gain (DCG)~\citep{Jarvelin2002-cm}.
Given relevance labels $\mathbf{y}\in\mathbb{R}_{\ge 0}^{M}$ and predicted scores $\mathbf{s}\in\mathbb{R}^{M}$, we define gains $g_i = 2^{y_i}-1$.
Let $\pi(\mathbf{s})$ denote the permutation sorting $\mathbf{s}$, so that
$s_{\pi_1(\mathbf{s})} \ge s_{\pi_2(\mathbf{s})} \ge \cdots \ge s_{\pi_M(\mathbf{s})}$, where $\pi_j(\mathbf{s})$ gives the index of the item ranked at position $j$.
Then DCG@k evaluates the quality of the top-$k$ ranked items:
\begin{equation}
\mathrm{DCG}@k(\mathbf{y},\mathbf{s})
\;=\;
\sum_{j=1}^{k}
\frac{g_{\pi_j(\mathbf{s})}}{\log_2(1+j)},
\end{equation}
and NDCG@k normalizes it by the ideal DCG:
\begin{equation}
\mathrm{NDCG}@k(\mathbf{y},\mathbf{s})
\;=\;
\frac{\mathrm{DCG}@k(\mathbf{y},\mathbf{s})}{\mathrm{IDCG}@k(\mathbf{y})+\varepsilon},
\end{equation}
where $\mathrm{IDCG}@k(\mathbf{y})$ is obtained by sorting items by $\mathbf{y}$, yielding the best DCG@k for $\mathbf{y}$, and $\varepsilon>0$ is for numerical stability.
However, DCG/NDCG are non-differentiable due to sorting, so we use a differentiable surrogate.

\paragraph{PiRank (relaxed NDCG@k)}
We adopt PiRank's differentiable relaxation of NDCG@k~\citep{Swezey2021-fg}.
Let $\hat{\mathbf{P}}(\mathbf{s};\tau)\in\mathbb{R}^{M\times M}$ denote the NeuralSort soft permutation matrix (temperature $\tau>0$) that relaxes the permutation induced by sorting $\mathbf{s}$, and let $\hat{\mathbf{P}}_{1:k}$ be its first $k$ rows.
With gains $\mathbf{g}\in\mathbb{R}_{\ge 0}^{M}$ ($g_i = 2^{y_i}-1$), the relaxed DCG@k is
\begin{equation}
\widehat{\mathrm{DCG}}@k(\mathbf{y},\mathbf{s})
\;=\;
\sum_{j=1}^{k}
\frac{\big[\hat{\mathbf{P}}_{1:k}(\mathbf{s};\tau)\,\mathbf{g}\big]_j}{\log_2(1+j)}.
\end{equation}
Using the same $\mathrm{IDCG}@k(\mathbf{y})$ as above, we define
\begin{equation}
\begin{aligned}
\label{eq:planner_pirank}
\widehat{\mathrm{NDCG}}@k(\mathbf{y},\mathbf{s})
&=
\frac{\widehat{\mathrm{DCG}}@k(\mathbf{y},\mathbf{s})}{\mathrm{IDCG}@k(\mathbf{y})+\varepsilon},\\
\mathcal{L}_{\mathrm{PiRank}}
&=
1-\widehat{\mathrm{NDCG}}@k(\mathbf{y},\mathbf{s}).
\end{aligned}
\end{equation}
Minimizing $\mathcal{L}_{\mathrm{PiRank}}$ yields stable gradients through $\hat{\mathbf{P}}$ while targeting the top-ranked unmasking decisions.

\section{Ground-Truth-Guided Unmasking Order}
\label{sec:where_to_unmask}

In this section, we explore whether \emph{supervised data} can guide not only token prediction, but also the unmasking order itself. To do so, we adopt a \emph{controlled decoding} protocol that (i) unmasks exactly one position per step and (ii) fills it with a greedy token update.
This protocol provides a clean testbed where performance differences stem exclusively from the ordering signal.

\subsection{Text generation with MDLMs}
\label{subsec:order_decoding}

We study a controlled generation process to isolate \emph{where-to-unmask} and its impact on generation quality. Starting from a fully masked completion $\mathbf{x}_1=(m,\ldots,m)$ of length $L$, we perform exactly $L$ unmasking steps, each unmasking one position, until no masked tokens remain.\footnote{For notational simplicity, we omit the prompt $\mathbf{c}$ from $\mu_\theta(\cdot\mid \mathbf{x},\mathbf{c})$ when the meaning is clear.}\,\footnote{This one-token-per-step protocol is equivalent to discretizing a linear masking/unmasking schedule (e.g., $\alpha_t = 1-t$) into $L$ reverse steps, so that the number of unmasked tokens increases by one at each step.}

At step $k$, let $M_k=\{i\in[L]:\mathbf{x}_k^{\,i}=m\}$ denote the set of masked positions.
Given an order score $s_i(\mathbf{x}_k)$
defined on masked positions, the next position to unmask is chosen as
\begin{equation}
\label{eq:select_topBk}
i^\ast \leftarrow \arg\max_{i\in M_k} s_i(\mathbf{x}_k).
\end{equation}
We then greedily fill the selected position:
\begin{equation}
\hat{x}^{\,i^\ast}\leftarrow \arg\max_{v\in\mathcal{V}}\mu_{\theta,i^\ast}(v\mid \mathbf{x}_k),
\label{eq:what_argmax}
\end{equation}
and keep all other positions unchanged, yielding next state $\mathbf{x}_{k-1}$. After $L$ steps, we obtain $\mathbf{x}_0$.

Algorithm~\ref{alg:order_based_decoding} summarizes this protocol.
Under this controlled setting, \emph{the only varying component across methods is the order score $s$ in Eq.~\eqref{eq:select_topBk}},
so any performance difference is attributable to \emph{where-to-unmask}.

\subsection{Order scores~\cite{Ye2025-ye, Kim2025-ge}}
\label{subsec:order_scores}

At inference time, order scores are typically computed from model confidence under the current masked state $\mathbf{x}_t$. Two standard choices are: (i) the maximum predicted probability at position $i$
(\texttt{Top-Prob})~\cite{Ye2025-ye}
\begin{equation}
s_i^{\mathrm{Top\text{-}Prob}}(\mathbf{x})
:=\max_{v\in\mathcal{V}}\mu_{\theta}^i(v\mid \mathbf{x}),
\label{eq:score_top}
\end{equation}
and (ii) the gap between the top two candidates (\texttt{Margin})~\cite{Kim2025-ge}
\begin{equation}
    s_i^{\mathrm{Margin}}(\mathbf{x})
    :=
    \mu_{\theta}^i(v^\star\mid \mathbf{x})
    -
    \max_{v\in\mathcal{V}\setminus\{v^\star\}}\mu_{\theta}^i(v\mid \mathbf{x}),
    \label{eq:score_margin}
\end{equation}
where $v^\star := \argmax_{v\in\mathcal{V}}\mu_{\theta}^i(v\mid \mathbf{x})$.

\begin{algorithm}[t]
    \caption{Controlled MDLM decoding.}
    \label{alg:order_based_decoding}
    {\small
    \begin{algorithmic}[1]
    \STATE \textbf{Input:} MDLM $\mu_\theta(\cdot\mid \mathbf{x},\mathbf{c})$, completion length $L$, prompt $\mathbf{c}$
    \STATE Initialize $\mathbf{x}_{L}\leftarrow (m,\ldots,m)$
    \FOR{$k=L,\ldots,1$}
    \STATE $M_k\leftarrow \{i\in[L]:\mathbf{x}_k^{\,i}=m\}$
    \STATE \textbf{Where-to-unmask:} compute $s_i(\mathbf{x}_k)$ for each $i\in M_k$
    \STATE \hspace{1em} Select position $i^\ast \leftarrow \arg\max_{i\in M_k} s_i(\mathbf{x}_k)$
    \STATE \textbf{What-to-unmask (greedy):}
    \STATE \hspace{1em} $\hat{x}^{\,i^\ast}\leftarrow \arg\max_{v\in\mathcal{V}}\mu_{\theta}^i(v\mid \mathbf{x}_k)$
    \STATE \textbf{Update:}
    \STATE \hspace{1em} $\mathbf{x}_{k-1}^{\,i}\leftarrow 
    \begin{cases}
    \hat{x}^{\,i^\ast} & \text{if } i=i^\ast,\\
    \mathbf{x}_{k}^{\,i} & \text{otherwise.}
    \end{cases}$
    \ENDFOR
    \STATE \textbf{Output:} completion $\mathbf{x}_{0}$
    \end{algorithmic}
    }
\end{algorithm}

\subsection{Ground-truth-guided order oracles (ours)}
\label{subsec:order_oracle}

Prior work largely relies on inference-time heuristics to decide the unmasking order~\cite{Chang2022-xf,Koh2025-qy,kim2025train}.
With supervised data, however, one may ask: \emph{can we use the ground-truth completion $\mathbf{x}_0$ as a teacher signal to search for better unmasking orders?}

To this end, we define two ground-truth-guided scores, \texttt{Gt-Prob} and \texttt{Gt-Margin}, as supervised counterparts to Top-Prob and Margin.
These scores depend on the ground-truth completion $\mathbf{x}_0$ and are therefore unavailable at inference. In this section, we use them as \emph{oracles} to quantify the potential improvement offered by supervised ordering.
In Sec.~\ref{sec:planner}, we leverage these oracle-derived orderings as supervision to learn a practical \emph{where-to-unmask} planner.

\paragraph{Gt-Prob}
We first score each masked position by the model probability assigned to the
ground-truth token:
\begin{equation}
s_i^{\mathrm{Gt\text{-}Prob}}(\mathbf{x}_t)
:=\mu_{\theta}^i(x_0^{\,i}\mid \mathbf{x}_t).
\label{eq:score_gtlik}
\end{equation}

\paragraph{Gt-Margin}
Because our controlled decoding in Eq.~\eqref{eq:what_argmax} commits to the \emph{argmax} token, a position is safe only when the ground-truth token \emph{dominates} competing alternatives. We therefore introduce \texttt{Gt-Margin}:
\begin{equation}
s_i^{\mathrm{Gt\text{-}Margin}}(\mathbf{x}_t)=
\mu_{\theta}^i(x_0^{\,i}\mid \mathbf{x}_t)
-\max_{v\neq x_0^{\,i}}\mu_{\theta}^i(v\mid \mathbf{x}_t).
\label{eq:score_gtmargin}
\end{equation}
Large positive margins indicate that greedy decoding is likely to place the correct token at $i$
if unmasked now, while small/negative margins signal ambiguity and suggest postponing $i$
until more context is unmasked.

\subsection{Evaluation}
\label{subsec:eval_setup}

\paragraph{Datasets.}
We evaluate on four benchmarks where the unmasking order is expected to substantially affect the generation trajectory: GSM8K~\cite{Cobbe2021-le}, MATH~\cite{Hendrycks2021-pe}, Sudoku 9$\times$9~\citep{david2020-sd}, and StrategyQA~\cite{Geva2021-zy}.
We measure performance using exact-match accuracy on all datasets.

\begin{figure*}[t]
    \centering
    \includegraphics[width=\linewidth]{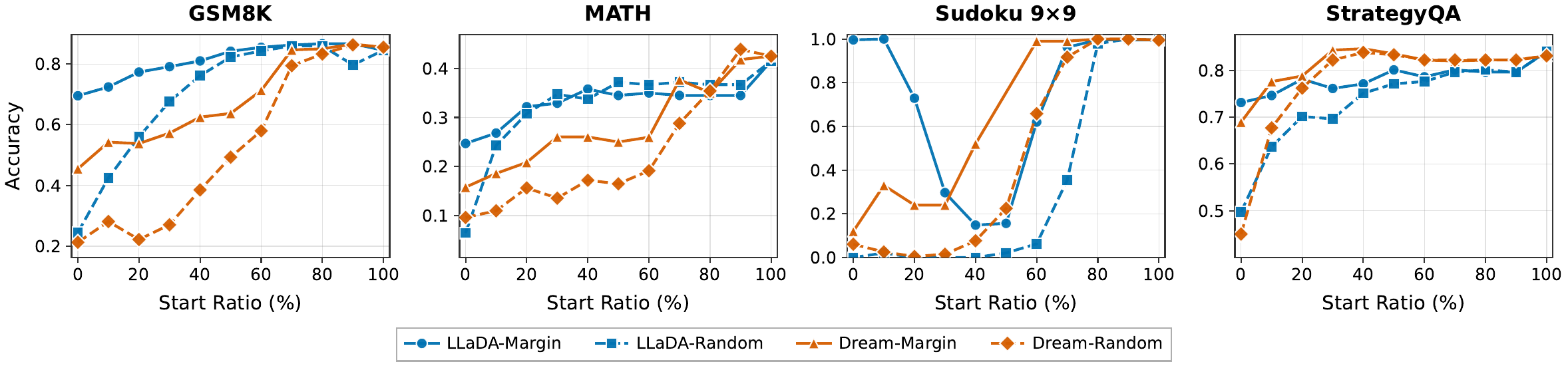}
    \caption{When is Gt-Margin important? We start from Gt-Margin ordering and replace it \emph{only within a single 10\% unmasking-step range} (0--10\%, 10--20\%, \ldots, 90--100\%) by either Margin or Random, keeping Gt-Margin elsewhere (i.e., 90\% of steps still use Gt-Margin). Across both LLaDA and Dream, perturbations in early steps lead to markedly worse final accuracy on most datasets, while late-step perturbations are less harmful, highlighting the outsized impact of early ordering decisions.}
    \label{fig:step_by_step_unmasking}
\end{figure*}

\paragraph{Models and controlled decoding protocol.}
We use LLaDA-8B~\cite{Nie2025-ne} and Dream-7B~\cite{Ye2025-pr} as base MDLMs, fine-tuned with LoRA~\citep{Hu2022-nl} on each dataset (Appendix~\ref{app:dataset_details}). At evaluation time, we run the controlled decoding procedure in Sec.~\ref{subsec:order_decoding}: (i) we unmask exactly one position per step (Eq.~\eqref{eq:select_topBk}), and (ii) token updates use greedy argmax (Eq.~\eqref{eq:what_argmax}). We vary \emph{only} the order score $s$ that determines the selected position.

\paragraph{Completion length.}
We use a dataset-specific fixed completion length $L$. For each dataset, we choose $L$ based on the empirical distribution of completion lengths, then keep it fixed across all examples and ordering strategies. Specifically, we use $L=256$ for GSM8K, $L=512$ for MATH, $L=128$ for Sudoku, and $L=128$ for StrategyQA. Details are provided in Appendix~\ref{app:completion_length_distribution}.

\paragraph{Order score baselines.}
We compare several baseline ordering strategies:
\texttt{Random} selects positions uniformly at random;
\texttt{AR} (autoregressive) unmasks left-to-right;
\texttt{Inverse-AR} unmasks right-to-left;
Top-Prob (Eq.~\eqref{eq:score_top}) and Margin (Eq.~\eqref{eq:score_margin}) are inference-time heuristics;
Gt-Prob and Gt-Margin (Eqs.~\eqref{eq:score_gtlik}--\eqref{eq:score_gtmargin}) are oracle baselines leveraging ground-truth completions.

\subsection{Results and Analysis}
\label{subsec:results_order_oracle}

Table~\ref{tab:fixlen} compares ordering strategies and shows that ground-truth-guided ordering signals can substantially improve generation under the same controlled decoding protocol.
Top-Prob and Margin outperform non-adaptive baselines (Random, AR, Inverse-AR), confirming that \emph{state-dependent} ordering matters. However, Gt-Prob and especially Gt-Margin yield much stronger performance across most datasets.
For instance, on GSM8K with LLaDA-8B, Margin reaches 0.605 while Gt-Margin achieves 0.845, showing that oracle-guided ordering alone can dramatically improve outcomes. On Sudoku, Gt-Margin nearly solves the task (0.995) while inference-time heuristics struggle.

\begin{table}[h]
    \centering
    \scriptsize
    \caption{Accuracy on GSM8K, MATH, Sudoku, and StrategyQA with different order scores.}
    \label{tab:fixlen}
    \begin{tabular}{lccccc}
        \toprule
        Model & Order score & GSM8K & MATH & Sudoku & StrategyQA \\
        \midrule
        \multirow{7}{*}{LLaDA-8B} 
        & Random      & 0.200 & 0.135 & 0.000 & 0.415 \\
        & AR          & 0.505 & 0.215 & 0.000 & 0.635 \\
        & Inverse-AR  & 0.266 & 0.065 & 0.000 & 0.635 \\
        & Top-Prob    & 0.600 & 0.215 & 0.105 & 0.635 \\
        & Margin      & 0.605 & 0.180 & 0.110 & 0.655 \\
        \cmidrule{2-6}
        & Gt-Prob     & 0.630 & 0.367 & 0.987 & 0.505 \\
        & Gt-Margin   & \textbf{0.845} & \textbf{0.415} & \textbf{0.995} & \textbf{0.840} \\
        \midrule
        \multirow{7}{*}{Dream-7B} 
        & Random      & 0.185 & 0.085 & 0.000 & 0.640 \\
        & AR          & 0.185 & 0.335 & 0.000 & 0.609 \\
        & Inverse-AR  & 0.375 & 0.125 & 0.000 & 0.015 \\
        & Top-Prob    & 0.395 & 0.230 & 0.210 & 0.785 \\
        & Margin      & 0.390 & 0.170 & 0.110 & 0.765 \\
        \cmidrule{2-6}
        & Gt-Prob     & 0.605 & 0.335 & 0.990 & 0.705 \\
        & Gt-Margin   & \textbf{0.855} & \textbf{0.425} & \textbf{0.995} & \textbf{0.835} \\
        \bottomrule
    \end{tabular}
\end{table}

Across most settings, Gt-Margin also substantially improves over Gt-Prob. This difference is expected under our controlled decoding (Eq.~\eqref{eq:what_argmax}): Gt-Prob can assign a relatively high score to a position even when the ground-truth token is not the top-1 prediction, in which case greedy decoding will deterministically commit an error if unmasked too early.
In contrast, Gt-Margin explicitly measures how decisively the ground-truth token dominates its strongest competitor, prioritizing safe positions and delaying ambiguous ones. These results motivate learning a practical where-to-unmask policy from supervised signals, which we pursue next.

\subsection{When does Gt-Margin matter?}
\label{sec:when_gtmargin_matters}

We next study \emph{when} Gt-Margin-style ordering is most influential.
We partition the trajectory into ten equal-length step ranges (0--10\%, 10--20\%, \ldots, 90--100\%) and replace the ordering \emph{only within that range} by either Margin or Random, keeping Gt-Margin elsewhere.

Figure~\ref{fig:step_by_step_unmasking} shows that early-step perturbations consistently degrade final accuracy across most datasets and both base models, whereas late-step perturbations are less harmful. This indicates that early where-to-unmask decisions strongly constrain the subsequent trajectory: once incorrect tokens are committed early, later steps cannot fully recover.

An exception is LLaDA on Sudoku when substituting early steps with Margin, where the effect is limited. We speculate that Sudoku's strong structural constraints make heuristic confidence sufficiently informative for early choices. Nevertheless, replacing early steps with Random still causes clear drops, confirming that early decisions remain critical.

\begin{figure}[t]
    \centering
    \includegraphics[width=\linewidth]{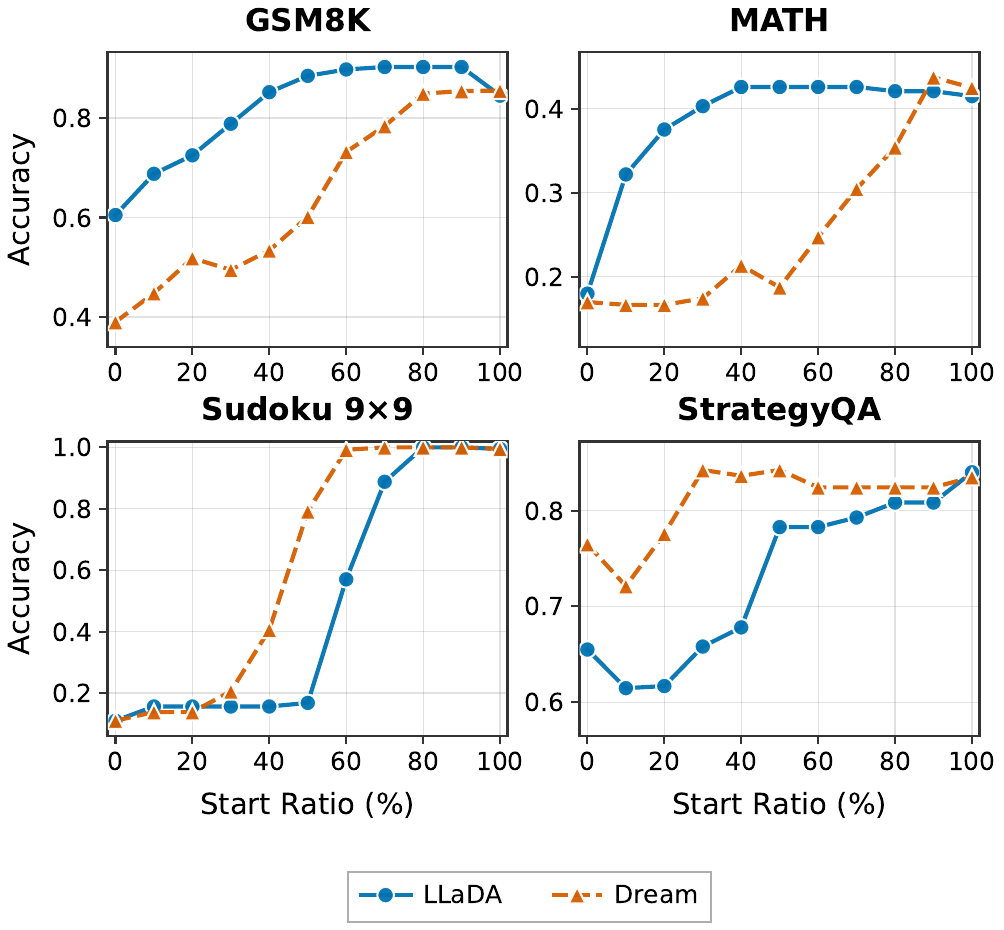}
    \caption{How long do we need oracle-like ordering? We decode with Gt-Margin for the first \(n\%\) of steps and switch to Margin for the remaining \(100-n\%\). Using Gt-Margin for approximately the first half is sufficient to recover most gains, indicating that oracle guidance is primarily beneficial early in decoding.}
    \label{fig:partial_plan_by_dataset}
\end{figure}

\subsection{How long do we need an oracle-like order?}
\label{sec:how_long_oracle_order}

Given the substantial impact of early ordering (Sec.~\ref{sec:when_gtmargin_matters}), we ask how long Gt-Margin-style guidance is needed to recover most gains over inference-time heuristics.
We construct a hybrid policy that uses Gt-Margin for the first $n\%$ of unmasking steps and then switches to Margin for the remaining $100-n\%$. We sweep $n\in\{0,10,\ldots,100\}$, where $n=0$ corresponds to pure Margin and $n=100$ to full Gt-Margin, while keeping other decoding components fixed.

Figure~\ref{fig:partial_plan_by_dataset} shows that applying Gt-Margin for approximately the first half of the trajectory is sufficient to attain near-Gt-Margin performance. This indicates that oracle-like ordering is most valuable early in decoding; once sufficient tokens are fixed, the remaining choices can be made effectively using inference-time heuristics such as Margin.

\subsection{Unmasking Order Comparison}
\begin{figure}[t]
    \centering
    \includegraphics[width=\linewidth]{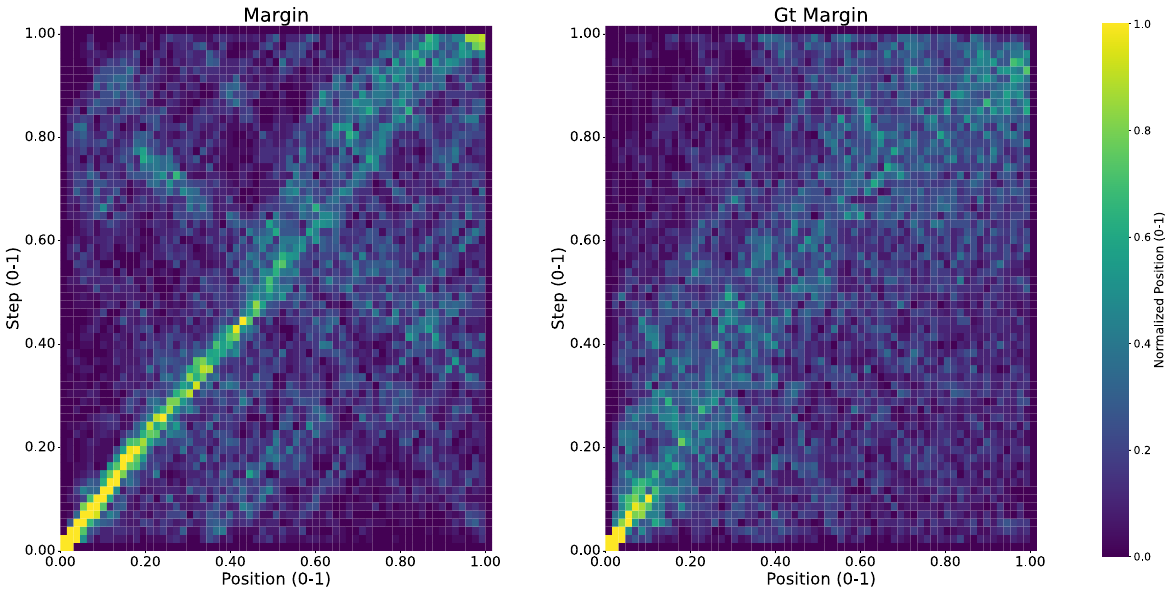}
    \caption{Empirical unmasking-order heatmaps on GSM8K. Both axes are normalized (\emph{x}: token position, \emph{y}: unmasking step), and color indicates how frequently a relative position is selected at a relative step. Margin concentrates tightly along the diagonal (near left-to-right behavior), while Gt-Margin retains a diagonal trend but places more mass off-diagonal, indicating more adaptive jumps to contextually easy positions.}
    \label{fig:unmasking_order_comparison}
\end{figure}

Figure~\ref{fig:unmasking_order_comparison} visualizes the empirical unmasking trajectories as a step-position heatmap on GSM8K.
Margin concentrates sharply along the diagonal, suggesting a nearly left-to-right schedule with limited deviations once an early prefix is favored.
In contrast, Gt-Margin exhibits a milder diagonal tendency with non-trivial off-diagonal weight, reflecting more frequent jumps to positions where the ground-truth token is well separated from alternatives.
Overall, Gt-Margin better exploits the flexible unmasking order of MDLMs rather than collapsing to a rigid schedule.

\section{Learning an Oracle-Guided Unmasking Ordering Planner}
\label{sec:planner}

\begin{figure*}[t]
    \centering
    \includegraphics[width=\linewidth]{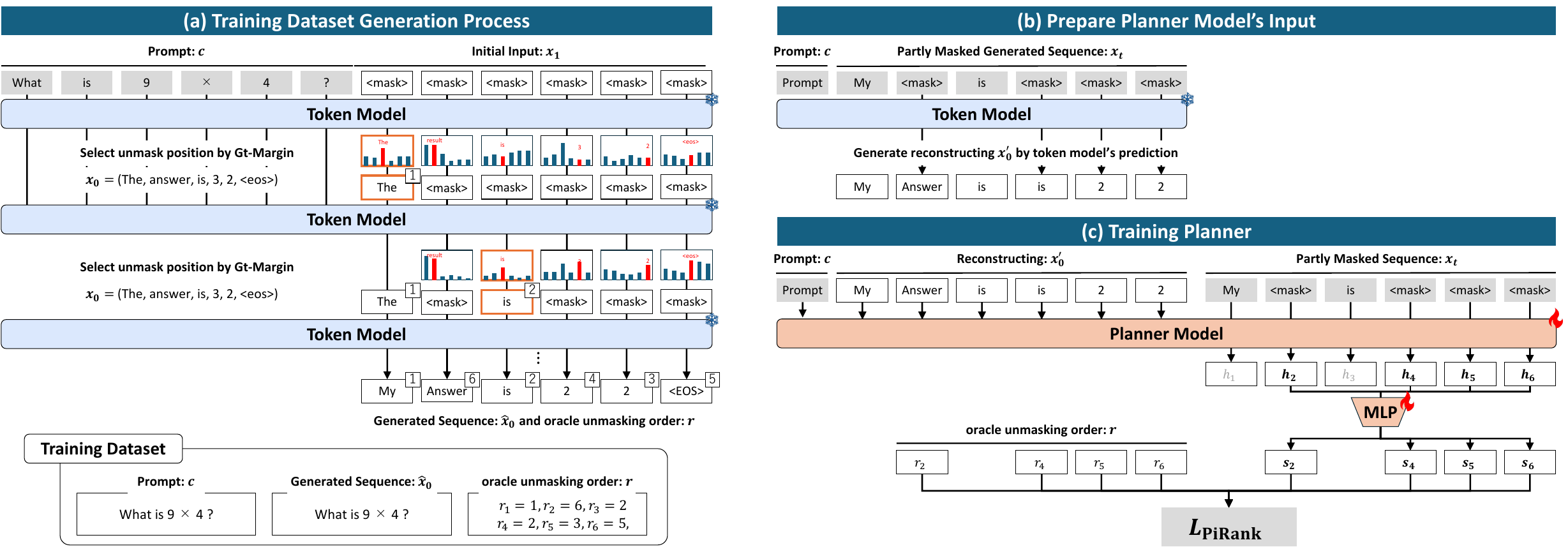}
    \caption{Overview of learning an oracle-guided unmasking ordering planner.
    \textbf{(a)} We run controlled decoding from a fully masked state and select the next position using Gt-Margin, yielding a generated completion $\hat{\mathbf{x}}_0$ and an oracle unmasking order $\mathbf{r}$.
    \textbf{(b)} We first sample a diffusion time $t$ to construct a partially masked sequence $\mathbf{x}_t$; we then form a reconstructed sequence $\mathbf{x}_0'$ by filling masked positions with the token model's argmax predictions and build the planner input as \texttt{prompt + $\mathbf{x}_0'$ + $\mathbf{x}_t$}.
    \textbf{(c)} The planner outputs priority scores $\mathbf{s}$ over masked positions via a MLP head and is trained with listwise ranking supervision to match $\mathbf{r}$.}
    \label{fig:method}
\end{figure*}

In this section, we explicitly separate two components:
the \emph{token model} $\mu_\theta$ for \emph{what-to-unmask} and a separately trained \emph{planner model} $f_\phi$ for \emph{where-to-unmask}.
The token model is the same MDLM used in Sec.~\ref{sec:where_to_unmask}; the planner is an additional module trained on top of it.

As shown in Figure~\ref{fig:method}, we train a supervised \emph{unmasking ordering planner} that predicts oracle-induced unmasking priorities from partially masked states.
Given a state $q=(\mathbf{c},\mathbf{x}_t)$ with masked set $M_t=\{i\in[L]:\mathbf{x}_t^{\,i}=m\}$, the planner outputs scores $\mathbf{s}\in\mathbb{R}^{|M_t|}$ over masked positions, where larger scores indicate earlier unmasking.
We formulate this as a learning-to-rank (LTR) problem.

\subsection{Oracle supervision from Gt-Margin}
\label{subsec:planner_teacher}

As shown in Figure~\ref{fig:method}(a), to supervise our planner model, we generate an \emph{oracle unmasking order} $\mathbf{r}$ by running the order-based decoding procedure described in Algorithm~\ref{alg:order_based_decoding}, using Gt-Margin as the scoring function.
This oracle-generation stage uses only the token model and ground-truth completions; the planner is then trained from these trajectories.

Specifically, given a training example consisting of a prompt $\mathbf{c}$ and its ground-truth completion $\mathbf{x}_0$, we initialize decoding from a fully masked sequence $\mathbf{x}_1=(m,\dots,m)$. At each decoding step, Gt-Margin (Eq.~\eqref{eq:score_gtmargin}) computes scores for all currently masked positions relative to the ground-truth completion $\mathbf{x}_0$. The highest-scoring position is selected and unmasked, gradually constructing a generated sequence $\hat{\mathbf{x}}_0$ and establishing an oracle-guided unmasking trajectory.

As Algorithm~\ref{alg:order_based_decoding} unmasks one position per step, the oracle trajectory is a rank vector $\mathbf{r}\in \{1,\dots,L\}^{L}$, where $r_i=k$ means position $i$ is unmasked at step $k$.

Executing this decoding procedure across the training set yields a dataset of tuples $(\mathbf{c}, \hat{\mathbf{x}}_0, \mathbf{r})$. The generated sequence $\hat{\mathbf{x}}_0$ ensures consistency with planner inputs during training, mitigating exposure bias. This dataset provides direct supervision for the planner to learn oracle-like unmasking orders from partially masked sequences.

\subsection{Training states: MDLM-style time sampling with rank-conditioned masking}
\label{subsec:planner_states}

A key challenge is distribution mismatch: at inference, the planner sees \emph{partially unmasked} states from reverse diffusion, not arbitrary random masks.
To match this regime while retaining standard MDLM training simplicity~\citep{Shi2024-mq,Sahoo2025-tm}, we construct training states via (i) \emph{time sampling} and (ii) \emph{rank-conditioned masking}.

We first sample a diffusion time $t\sim\mathrm{Unif}(0,1)$ to determine the unmasking budget $u=\lfloor \alpha_t L \rfloor$ via the schedule $\alpha_t$ (Sec.~\ref{sec:prelim}).
Instead of deterministically unmasking the top-$u$ ranks, we construct $\mathbf{x}_t$ stochastically: each position $i$ is independently unmasked with probability increasing with earlier oracle rank $r_i$ and larger budget $u$, yielding diverse oracle-consistent intermediate states (Appendix~\ref{app:planner_state_sampling}).

\subsection{Planner input and architecture}
\label{subsec:planner_input}

\paragraph{Input representation.}
As shown in Figure~\ref{fig:method}(b), the planner takes a prompt $\mathbf{c}$ and a partially masked state $\mathbf{x}_t$.
We augment the input \texttt{prompt + $\mathbf{x}_t$} with an auxiliary token prediction $\mathbf{x}_0'$ by filling masked positions with the token model's argmax predictions:
\begin{equation}
\mathbf{x}_0'^{\,i}=
\begin{cases}
\mathbf{x}_t^{\,i} & (\mathbf{x}_t^{\,i}\neq m),\\
\arg\max_{v\in\mathcal{V}}\mu_{\theta}^i(v\mid \mathbf{x}_t,\mathbf{c}) & (\mathbf{x}_t^{\,i}=m).
\end{cases}
\end{equation}
The planner input is then \texttt{prompt + $\mathbf{x}_0'$ + $\mathbf{x}_t$}.
Intuitively, $\mathbf{x}_t$ makes \emph{where uncertainty remains} explicit through its mask pattern, while $\mathbf{x}_0'$ provides a hypothesis of \emph{what the missing content might be}.
This reconstruct-then-plan input lets the planner judge which masked positions are safe to reveal.

\paragraph{Backbone and scoring head.}
As shown in Figure~\ref{fig:method}(c), we implement the planner as a bidirectional Transformer backbone with a lightweight scoring head.
Let $\mathbf{h}_i$ denote the final-layer hidden state for completion position $i$ in the $\mathbf{x}_t$ segment.
For each $i\in M_t$, we compute a scalar priority score via a three-layer MLP:
\begin{equation}
s_i
=
\mathrm{MLP}_3(\mathbf{h}_i),
\qquad i\in M_t.
\end{equation}
We then rank masked positions by sorting $\{s_i\}_{i\in M_t}$ in descending order.
This design keeps the planner's cost comparable to a single Transformer forward pass and allows seamless integration into standard MDLM sampling without modifying the underlying token model.

\subsection{Listwise LTR objective for the planner}
\label{subsec:planner_topk}

At each reverse step, the model unmasks only a small subset of positions (Sec.~\ref{sec:prelim}), making accurate top-$k$ ranking critical.
We formulate planner training as a listwise learning-to-rank problem: given the masked set $M_t$, the planner scores each position to induce an unmasking ranking.

We derive a target relevance distribution $\mathbf{y}$ from the oracle ranks $\mathbf{r}$, restricted to $M_t$ (Appendix~\ref{app:planner_ltr_objective}), assigning higher values to positions the oracle would unmask earlier.
As shown in Figure~\ref{fig:method}(c), we train the planner by minimizing the PiRank loss (relaxed NDCG@k, Eq.~\eqref{eq:planner_pirank}), comparing predicted scores $\mathbf{s}$ against target relevance $\mathbf{y}$.

\subsection{Inference: plug-and-play where-to-unmask}
\label{subsec:planner_inference}

At inference, we replace heuristic scores (e.g., Margin) with the planner's scores $s_i$ to select unmasking positions.
The planner controls only \emph{where-to-unmask}; token predictions remain governed by the original model $\mu_\theta$.

\subsection{Partial-plan decoding.}
\label{subsec:planner_partial_plan}
Motivated by Sec.~\ref{sec:how_long_oracle_order} (Fig.~\ref{fig:partial_plan_by_dataset}), which shows oracle-like ordering is most beneficial early in generation, we apply the learned planner only during the first half of the reverse trajectory and switch to the Margin heuristic for remaining steps.
This hybrid strategy focuses the planner on the high-uncertainty regime where oracle guidance is most valuable.

\subsection{Evaluation}
\label{subsec:planner_eval}

\paragraph{Evaluation setup}
We evaluate on the same four benchmarks as Sec.~\ref{subsec:eval_setup}: GSM8K~\cite{Cobbe2021-le}, MATH~\cite{Hendrycks2021-pe}, Sudoku 9$\times$9~\citep{david2020-sd}, and StrategyQA~\cite{Geva2021-zy}, using LLaDA-8B~\cite{Nie2025-ne}.
We train a separate planner (LLaDA-8B backbone with LoRA and MLP head) and vary only the \emph{where-to-unmask} strategy: Random, Margin, or our learned planner.
See Appendix~\ref{app:implementation_details} for details.

\subsection{Results and Analysis}
\label{subsec:planner_results}

Table~\ref{tab:planner_ablation} evaluates different ordering strategies under the same MDLM sampling procedure.
Our learned planner yields further gains on GSM8K and MATH (and modest improvements on StrategyQA), demonstrating that oracle-induced priorities from Gt-Margin can be distilled into a deployable where-to-unmask policy without modifying the token model.
On Sudoku, the planner does not yet match Margin, suggesting that distilling oracle-like ordering for highly structured completions remains challenging.

\begin{table}[t]
    \centering
    \scriptsize
    \caption{Accuracy on GSM8K, MATH, Sudoku, and StrategyQA with different order scores.}
    \label{tab:planner_ablation}
    \begin{tabular}{lcccc}
        \toprule
        Order score & GSM8K & MATH & Sudoku & StrategyQA \\
        \midrule
        Random & 0.200 & 0.135 & 0.000 & 0.415 \\
        Margin & 0.605 & 0.180 & \textbf{0.110} & 0.655 \\
        \cmidrule{1-5}
        Ours w/o partial plan & 0.635 & 0.180 & 0.045 & 0.645 \\
        Ours w/o token pred & 0.625 & 0.175 & 0.000 & 0.635 \\
        \cmidrule{1-5}
        Ours & \textbf{0.705} & \textbf{0.285} & 0.085 & \textbf{0.685} \\
        \bottomrule
    \end{tabular}
\end{table}

\paragraph{Ablations.}
We ablate two components:
\textbf{w/o partial plan} uses the planner throughout the trajectory instead of only early steps.
\textbf{w/o token pred} removes the auxiliary token hypothesis $\mathbf{x}_0'$ from the planner input.
These ablations isolate (i) the benefit of focusing oracle-like planning on the early trajectory (Sec.~\ref{sec:how_long_oracle_order}), and (ii) the value of token-level hypotheses for assessing which masked positions are safe to reveal.
Table~\ref{tab:planner_ablation} shows both ablations degrade performance, confirming the importance of early-stage planning and the reconstruct-then-plan input.

\section{Related Work}

\paragraph{Diffusion language models for text and masked diffusion LLMs.}
Diffusion models have been adapted to discrete sequences, enabling non-autoregressive generation via iterative denoising.
D3PMs and continuous-time CTMC formulations provide principled objectives and transition kernels \citep{Austin2021-pj,Campbell2022-xd,Lou2024-op}.
Masked diffusion gradually replaces tokens with \texttt{<mask>} and fills them in; simplified objectives and large-scale training have enabled competitive MDLMs \citep{Shi2024-mq,Sahoo2025-tm,Nie2025-ne,Ye2025-pr,Bie2025-ae}.
This approach supports flexible editing and infilling \citep{Stern2019-xw,Ghazvininejad2019-ec,Lee2025-qu,Zhang2025-cx,Chang2022-xf}, with conditional variants also proposed \citep{Koh2025-qy}.
Recent work studies training/sampling behavior \citep{Zheng2025-xt}, autoregressive-diffusion hybrids \citep{Arriola2025-qc}, and data-efficiency \citep{Ni2025-lf}.
For a broader overview, please refer to \citet{Li2025-rv}.

\paragraph{Ground-truth-guided supervised unmasking (ours).}
Prior work either selects unmasking orders using inference-time uncertainty heuristics \citep{Chang2022-xf,Koh2025-qy,kim2025train} or improves sampling via additional optimization/search such as RL-trained policies \citep{Hong2025-np}, lookahead/path selection \citep{Lee2025-up,Peng2025-oi}, or remasking-based refinement \citep{Wang2025-st}.
In contrast, we introduce Gt-Margin, a ground-truth-dependent oracle that ranks positions by the margin between correct and competing tokens, inducing an easy-to-hard order.
We distill this oracle into a planner trained via supervised learning-to-rank, yielding a deployable \emph{where-to-unmask} policy without modifying the token model or requiring on-policy rollouts.
Our analysis shows early steps matter most, motivating partial-plan decoding that applies the planner only in the high-mask regime.

\section{Conclusion}
We studied the \emph{where-to-unmask} decision in MDLMs, which is critical for generation quality but typically handled by inference-time heuristics.
We introduced Gt-Margin, a ground-truth-derived score that induces an oracle easy-to-hard unmasking order by measuring how decisively the model prefers the ground-truth token at each position.
We analyzed when ordering matters most, observed consistent gains on logical reasoning benchmarks, and trained a supervised planner to imitate the oracle ranking, improving decoding without modifying the token model.

\clearpage

\section*{Acknowledgments}
This work has been supported by the JST Moonshot Research and Development Program \mbox{JPMJMS2236-8}.

\bibliography{library}
\bibliographystyle{icml2026}

\clearpage
\appendix
\section{Planner Training Details}
\subsection{Stochastic rank-conditioned masking for planner training}
\label{app:planner_state_sampling}

This section details how we construct the partially masked training state $\mathbf{x}_t$ from an oracle rank vector $\mathbf{r}\in\{1,\ldots,L\}^L$.

\paragraph{Setup.}
Let $L$ be the completion length and $r_i$ the oracle unmasking rank for completion position $i\in[L]$ (smaller means earlier under the oracle).
Given a target unmasking budget $u\in\{0,1,\ldots,L\}$, our goal is to create a mask pattern that (i) preferentially keeps low-rank positions visible and (ii) remains stochastic, so the planner observes a variety of plausible intermediate states.

\paragraph{Time sampling}
Sec.~\ref{subsec:planner_states} describes time sampling as $t\sim\mathrm{Unif}(0,1)$ with an unmasking budget $u=\lfloor \alpha_t L\rfloor$ induced by the MDLM schedule $\alpha_t$.
In implementation it is convenient to sample a discrete index $\tilde{t}\in\{0,1,\ldots,L\}$ and set $u:=L-\tilde{t}$ (so $\tilde{t}=L$ is fully masked and $\tilde{t}=0$ is fully unmasked).

For partial plan training (Sec.~\ref{subsec:planner_partial_plan}), we restrict to the early phase of reverse diffusion by sampling $\tilde{t}\sim \mathrm{Unif}(\{\tilde{t}_{\min},\ldots,L\})$ where $\tilde{t}_{\min}=\lceil (1-\rho)L\rceil$ for an \texttt{initial\_unmask\_ratio} $\rho\in(0,1]$, corresponding to the first $\rho$ fraction of reverse steps. In practice, we use $\rho=0.5$.

\paragraph{Bernoulli masking with a soft rank threshold.}
For each completion position $i\in[L]$ with oracle rank $r_i$, we compute an unmask probability
\begin{equation}
p_i(\tilde{t})
\;=\;
\sigma\!\Big(\frac{u-r_i}{\tau}\Big)
\;=\;
\sigma\!\Big(\frac{(L-\tilde{t})-r_i}{\tau}\Big),
\end{equation}
where $\sigma(\cdot)$ is the logistic sigmoid and $\tau>0$ is a temperature.
We then sample independent Bernoulli variables $z_i\sim\mathrm{Bernoulli}(p_i(\tilde{t}))$ and set
\begin{equation}
\mathbf{x}_t^{\,i}=
\begin{cases}
\mathbf{x}_0^{\,i} & (z_i=1),\\
m & (z_i=0).
\end{cases}
\end{equation}
This yields a stochastic mask pattern with $\mathbb{E}\big[\sum_i z_i\big]=\sum_i p_i(\tilde{t})$, and concentrates unmasking probability on small-rank (easy/early) positions.
The temperature $\tau$ controls the sharpness of the transition around the soft threshold $r_i\approx u$: as $\tau\to 0$, masking becomes deterministic (hard-thresholding at $r_i=u$), while larger $\tau$ yields more stochastic mask patterns.
In practice we set $\tau=5$ to balance fidelity to the oracle rank ordering with sufficient diversity in training states.

\paragraph{Sampling procedure.}
Algorithm~\ref{alg:planner_state_sampling} summarizes the stochastic rank-conditioned masking used to form $\mathbf{x}_t$.
\begin{algorithm}[t]
    \caption{Stochastic rank-conditioned masking.}
    \label{alg:planner_state_sampling}
    {\small
    \begin{algorithmic}[1]
    \STATE \textbf{Input:} completion tokens $\mathbf{x}_0$, ranks $\mathbf{r}$, length $L$, mask token $m$, temperature $\tau$
    \STATE Sample $\tilde{t}\in\{0,\ldots,L\}$ (optionally restricted); set $u\leftarrow L-\tilde{t}$
    \FOR{$i=1,\ldots,L$}
        \STATE $p_i \leftarrow \sigma\big((u-r_i)/\tau\big)$
        \STATE Sample $z_i\sim\mathrm{Bernoulli}(p_i)$
        \STATE $\mathbf{x}_t^{\,i} \leftarrow \mathbf{x}_0^{\,i}$ if $z_i=1$, else $\mathbf{x}_t^{\,i}\leftarrow m$
    \ENDFOR
    \STATE \textbf{Output:} partially masked state $\mathbf{x}_t$
    \end{algorithmic}
    }
\end{algorithm}

\subsection{Target relevance distribution for listwise LTR}
\label{app:planner_ltr_objective}

This section details how we construct the masked-only target relevance distribution $\mathbf{y}$ used for listwise planner training (Sec.~\ref{subsec:planner_topk}).

\paragraph{Setup.}
For a training state with masked set $M_t$ and oracle rank vector $\mathbf{r}\in\{1,\ldots,L\}^L$, we restrict ranks to the masked positions and write $\mathbf{r}_{M_t}:=\{r_i\}_{i\in M_t}\in\{1,\ldots,L\}^{M}$ where $M=|M_t|$.

\paragraph{Top-$k$ support.}
We clip the cutoff via $k\leftarrow\min(k,M)$ and define $\mathcal{P}_t\subseteq M_t$ as the set of $k$ masked positions with the \emph{smallest} ranks (earliest under the oracle).
Let $r_{\max}=\max_{i\in\mathcal{P}_t} r_i$ be the largest rank among the selected positions.

\paragraph{Target relevance distribution.}
We first assign an unnormalized target relevance score $\ell_i$ to each masked position, and then normalize by a softmax over the masked set:
\begin{equation}
\label{eq:planner_relevance}
\ell_i
\;=\;
\begin{cases}
\frac{r_{\max}-r_i}{\tau_{\mathrm{tgt}}} & (i\in\mathcal{P}_t),\\
-\infty & (i\in M_t\setminus\mathcal{P}_t),
\end{cases}
\qquad
\mathbf{y}=\mathrm{softmax}(\boldsymbol{\ell}),
\end{equation}
where $\tau_{\mathrm{tgt}}>0$ is a temperature.
Because $\ell_i=-\infty$ outside $\mathcal{P}_t$, we have $y_i=0$ for non-prioritized masked positions, while earlier-ranked positions within $\mathcal{P}_t$ receive larger relevance due to the term $r_{\max}-r_i$.
The temperature $\tau_{\mathrm{tgt}}$ controls how concentrated $\mathbf{y}$ is among the oracle top-$k$ set; we use $\tau_{\mathrm{tgt}}=5$ in practice.

\section{Implementation Details}
\label{app:implementation_details}

Unless otherwise noted, we use the same optimization hyperparameters for training both the token model and the planner model:
\begin{table}[h]
\centering
\small
\begin{tabular}{ll}
\toprule
\textbf{Hyperparameter} & \textbf{Value} \\
\midrule
\texttt{lr} & $2\times 10^{-4}$ \\
\texttt{weight\_decay} & $1\times 10^{-1}$ \\
\texttt{adam\_beta1} & $0.9$ \\
\texttt{adam\_beta2} & $0.95$ \\
\texttt{max\_grad\_norm} & $1.0$ \\
\texttt{lr\_scheduler\_type} & \texttt{cosine} \\
\texttt{warmup\_ratio} & $0.1$ \\
\texttt{per\_device\_train\_batch\_size} & $2$ \\
\texttt{gradient\_accumulation\_steps} & $4$ \\
\bottomrule
\end{tabular}
\end{table}

For the MLP head, we use a 3-layer MLP with GELU activations, where the hidden widths are set to \texttt{intermediate\_size\_1} $=$ \texttt{hidden\_size}$/2$ and \texttt{intermediate\_size\_2} $=$ \texttt{hidden\_size}$/4$, and the final layer outputs a scalar.
For LLaDA, \texttt{hidden\_size} (i.e., \texttt{hidden\_dim}) is 4096.

To mitigate overfitting, we train on Sudoku for 2 epochs, and for all other datasets we train for 5 epochs (for both planner and token model training).
All experiments are run on a single NVIDIA A100 GPU.
For the PiRank loss (relaxed NDCG@k), we use $k=30$ for all datasets.

For parameter-efficient fine-tuning, we apply LoRA~\citep{Hu2022-nl} to update the base model parameters for both the token model and the planner model. For the planner model, the MLP head is additionally trained with full parameters.

\section{Dataset Details}
\label{app:dataset_details}

\subsection{Completion Length Distribution and Selection}
\label{app:completion_length_distribution}

\begin{figure*}[t]
    \centering
    \includegraphics[width=\textwidth]{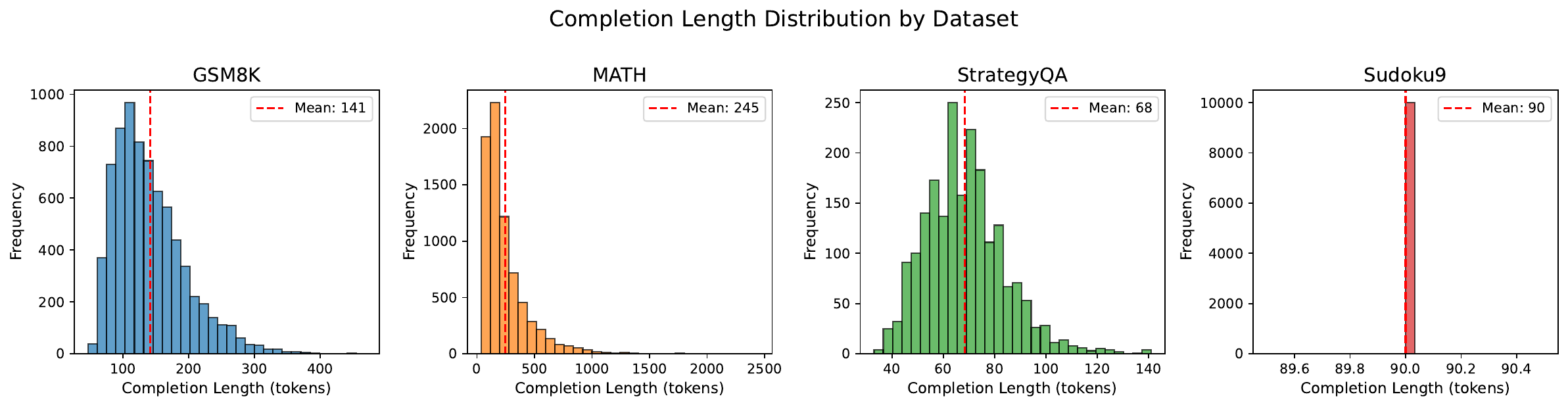}
    \caption{Completion length distributions (in tokens) for each dataset. The red dashed line indicates the mean completion length. GSM8K, MATH, and StrategyQA show varying degrees of right-skewed distributions, while Sudoku has a fixed length of 90 tokens due to the deterministic output format.}
    \label{fig:completion_length_distribution}
\end{figure*}

Figure~\ref{fig:completion_length_distribution} shows the distribution of completion lengths (in tokens) across our four evaluation datasets.
Each dataset exhibits distinct characteristics:
\begin{itemize}
    \item \textbf{GSM8K}: Grade-school math word problems with a mean completion length of 141 tokens. The distribution is right-skewed with most completions between 50--200 tokens.
    \item \textbf{MATH}: Competition-level mathematics problems with a mean completion length of 245 tokens. This dataset shows the highest variance, with a long tail extending beyond 2000 tokens for complex proofs.
    \item \textbf{StrategyQA}: Commonsense reasoning questions with a mean completion length of 68 tokens. Completions are relatively short and concentrated around the mean.
    \item \textbf{Sudoku}: 9$\times$9 Sudoku puzzles with a fixed completion length of exactly 90 tokens (81 digits for the solution plus formatting tokens).
\end{itemize}

Based on these distributions, we set the maximum completion length $L$ for each dataset as follows: $L=128$ for GSM8K, $L=512$ for MATH, $L=128$ for Sudoku, and $L=128$ for StrategyQA.
For GSM8K and StrategyQA, $L=128$ covers the majority of completions.
For MATH, we use $L=512$ to accommodate the longer reasoning chains required for complex mathematical proofs.
For Sudoku, although the actual completion length is fixed at 90 tokens, we use $L=128$ to provide a small buffer.

\subsection{Prompt Templates and Dataset Examples}
\label{app:dataset_examples}

We use task-specific prompt templates to elicit structured responses from the model.

For \textbf{GSM8K} and \textbf{MATH}, we use a common template that requests both reasoning steps and a final answer:
\begin{promptbox}{Prompt Template (GSM8K / MATH)}
\begin{lstlisting}[style=prompttemplate]
Respond in the following format:
<reasoning>
...
</reasoning>
<answer>
...
</answer>

{question}
\end{lstlisting}
\end{promptbox}

For \textbf{Sudoku}, we provide detailed instructions about the puzzle format and rules, requesting only the final solution without intermediate reasoning:
\begin{promptbox}{Prompt Template (Sudoku)}
\begin{lstlisting}[style=prompttemplate]
Please solve the following 9x9 Sudoku puzzle. The puzzle is
provided as an 81-character string reading left-to-right,
top-to-bottom, where '0' represents empty cells.

Rules:
- Fill empty cells with digits 1-9
- Each row must contain digits 1-9 exactly once
- Each column must contain digits 1-9 exactly once
- Each 3x3 box must contain digits 1-9 exactly once

Important: Your solution must be a COMPLETE 81-character string
with only the digits 1-9, representing your final solved grid.

Respond ONLY with the final solution enclosed in <answer> tags.
Do not include any reasoning.
<answer>
[81-character solution string with no spaces or separators]
</answer>

Solve the following Sudoku puzzle:
{sudoku_puzzle}
\end{lstlisting}
\end{promptbox}

For \textbf{StrategyQA}, we use a template similar to GSM8K/MATH but constrain the answer to be a boolean value:
\begin{promptbox}{Prompt Template (StrategyQA)}
\begin{lstlisting}[style=prompttemplate]
Respond in the following format:
<reasoning>
...
</reasoning>
<answer>
Only provide "true" or "false".
</answer>

{question}
\end{lstlisting}
\end{promptbox}

Tables~\ref{tab:example_gsm8k}--\ref{tab:example_strategyqa} show concrete examples of the question portion of each prompt (after the format instruction) and the corresponding expected completion.

\begin{table*}[t]
\centering
\caption{Example from GSM8K dataset.}
\label{tab:example_gsm8k}
\small
\begin{tabular}{p{0.15\textwidth}p{0.80\textwidth}}
\toprule
\textbf{Question} & Natalia sold clips to 48 of her friends in April, and then she sold half as many clips in May. How many clips did Natalia sell altogether in April and May? \\
\midrule
\textbf{Completion} & \texttt{<reasoning>} \\
& Natalia sold 48/2 = \texttt{<<48/2=24>>}24 clips in May. \\
& Natalia sold 48+24 = \texttt{<<48+24=72>>}72 clips altogether in April and May. \\
& \texttt{</reasoning>} \\
& \texttt{<answer>} \\
& 72 \\
& \texttt{</answer>} \\
\bottomrule
\end{tabular}
\end{table*}

\begin{table*}[t]
\centering
\caption{Example from MATH dataset.}
\label{tab:example_math}
\small
\begin{tabular}{p{0.15\textwidth}p{0.80\textwidth}}
\toprule
\textbf{Question} & When Claire divides her cupcakes into groups of 5, she has 3 remaining, and when she divides her cupcakes into groups of 7, she has 4 remaining. If Claire has fewer than 60 cupcakes, what is the sum of all possible quantities of cupcakes that she could have? \\
\midrule
\textbf{Completion} & \texttt{<reasoning>} \\
& If Claire has $N$ cupcakes, we know that $N = 5x+3$ and $N = 7y+4$ for some integers $x$ and $y$. \\
& Equating these two forms of $N$, we have $7y+1=5x$. \\
& We also know that $N<60$. \\
& We can write out all the sufficiently small possible values of $7y+1$: $1,8,15,22,29,36,43,50,57$. \\
& Of these, the numbers that are also of the form $5x$ are $15$ and $50$. \\
& These correspond to solutions $N=18,$ $N=53$. \\
& Thus, the sum of all possible quantities of cupcakes is $53+18 = \boxed{71}$. \\
& \texttt{</reasoning>} \\
& \texttt{<answer>} \\
& 71 \\
& \texttt{</answer>} \\
\bottomrule
\end{tabular}
\end{table*}

\begin{table*}[t]
\centering
\caption{Example from Sudoku dataset.}
\label{tab:example_sudoku}
\small
\begin{tabular}{p{0.15\textwidth}p{0.80\textwidth}}
\toprule
\textbf{Puzzle} & 100503700603008090000009800010000000876100000000006000000000007080907604700060312 \\
\midrule
\textbf{Completion} & \texttt{<answer>} \\
& 198543726643278591527619843914735268876192435235486179462351987381927654759864312 \\
& \texttt{</answer>} \\
\bottomrule
\end{tabular}
\end{table*}

\begin{table*}[t]
\centering
\caption{Example from StrategyQA dataset.}
\label{tab:example_strategyqa}
\small
\begin{tabular}{p{0.15\textwidth}p{0.80\textwidth}}
\toprule
\textbf{Question} & Would the average Hawaiian male experience more days on Earth compared to a wild cane toad? \\
\midrule
\textbf{Completion} & \texttt{<reasoning>} \\
& The average lifespan of a male born in Hawaii is 79.3 years. \\
& Cane toads have a life expectancy of 10 to 15 years in the wild. \\
& 79.3 years is greater than 15 years, so the average Hawaiian male would experience more days on Earth than a wild cane toad. \\
& \texttt{</reasoning>} \\
& \texttt{<answer>} \\
& true \\
& \texttt{</answer>} \\
\bottomrule
\end{tabular}
\end{table*}

\end{document}